# Transformer-Based Bearing Fault Detection using Temporal Decomposition Attention Mechanism

Marzieh Mirzaeibonehkhater *, Mohammad Ali Labbaf-Khaniki**, Mohammad Manthouri***

* Department of Electrical and Computer Engineering Indiana University-Purdue University

**Faculty of Electrical Engineering, K.N. Toosi University of Technology, Tehran, Iran

***Faculty of Electrical and Electronic Engineering Department, Shahed university

*marzieh89mirzaei@gmail.com

**mohamad95labafkh@gmail.com

***mmanthouri@shahed.ac.ir

**Abstract:**

Bearing fault detection is a critical task in predictive maintenance, where accurate and timely fault identification can prevent costly downtime and equipment damage. Traditional attention mechanisms in Transformer neural networks often struggle to capture the complex temporal patterns in bearing vibration data, leading to suboptimal performance. To address this limitation, we propose a novel attention mechanism, Temporal Decomposition Attention (TDA), which combines temporal bias encoding with seasonal-trend decomposition to capture both long-term dependencies and periodic fluctuations in time series data. Additionally, we incorporate the Hull Exponential Moving Average (HEMA) for feature extraction, enabling the model to effectively capture meaningful characteristics from the data while reducing noise. Our approach integrates TDA into the Transformer architecture, allowing the model to focus separately on the trend and seasonal components of the data. Experimental results on the Case Western Reserve University (CWRU) bearing fault detection dataset demonstrate that our approach outperforms traditional attention mechanisms and achieves state-of-the-art performance in terms of accuracy and interpretability. The HEMA-Transformer-TDA model achieves an accuracy of 98.1%, with exceptional precision, recall, and F1-scores, demonstrating its effectiveness in bearing fault detection and its potential for application in other time series tasks with seasonal patterns or trends.

**Keywords.** Bearing Fault Detection, Feature Extraction, Deep Learning, Transformer, Attention Mechanism.

## 1) Introduction

Industrial systems and plants are the backbone of modern economies, providing essential goods and services to societies worldwide. However, these complex systems are prone to faults and failures, which can have catastrophic consequences, including equipment damage, production downtime, and even loss of life [1]. The timely detection of faults is crucial to prevent such disasters and ensure the reliability, efficiency, and safety of industrial operations. In fact, studies have shown that unexpected equipment failures can result in losses of up to 20% of total production capacity, highlighting the need for effective fault detection and diagnosis strategies. Moreover, the increasing complexity of modern industrial systems, coupled with the growing demand for productivity and efficiency, has created a pressing need for advanced fault detection techniques [2]. In this context, the development of intelligent fault detection systems that can learn from data and adapt to changing operating conditions has become a critical research area, with significant implications for the reliability, safety, and profitability of industrial operations.

Feature extraction is a vital process in fault detection for time series data, as it enables the identification of meaningful patterns and anomalies that may indicate potential issues. Moving averages are particularly important in this context, as they smooth the data to remove random noise while preserving the underlying trends and periodic behaviors. By highlighting shifts or deviations in these patterns, moving averages provide valuable insights into the operational dynamics of a system [3]. This is especially critical in fault detection, where capturing subtle changes can be the difference between early diagnosis and system failure. Moving averages facilitate the preprocessing of time series data, enhancing the detection of anomalies and supporting the development of more accurate and reliable fault detection systems [4].

Deep learning methods, such as convolutional neural networks (CNNs) and recurrent neural networks (RNNs), have shown great promise in fault detection and diagnosis in industrial systems, due to their ability to learn complex patterns and relationships in data [5]. However, conventional deep learning methods like CNNs and RNNs face key challenges in fault detection for industrial systems, particularly when dealing with time series data [6]. Specifically, CNNs struggle with capturing long-term temporal dependencies, while RNNs suffer from vanishing gradients, limiting their ability to learn long-range patterns. Furthermore, these methods often require extensive feature engineering and may not effectively handle the non-stationarity and seasonality commonly present in industrial systems [7].

In contrast to traditional deep learning methods, Transformer-based architectures have demonstrated exceptional proficiency in fault detection tasks, particularly when dealing with complex time series data [8]. Transformers, a neural network architecture, have gained significant attention for their ability to handle sequential data in both Natural Language Processing (NLP) and time series analysis. Their self-attention mechanism enables them to capture complex, long-range dependencies within data sequences, making them highly effective for tasks involving intricate temporal patterns. The ability of Transformers to learn contextual relationships between different parts of the data and to weigh the importance of different input elements makes them particularly well-suited for fault detection tasks, where the relationships between different variables can be complex and nuanced [9]. Additionally, the parallelization capabilities of Transformers enable faster training and inference times, making them a more efficient choice for large-scale industrial applications. Furthermore, the interpretability of Transformers, through attention weights, provides valuable insights into the decision-making process, allowing for better understanding and trust in the model's predictions [10].

The attention mechanism is a crucial component of the Transformer architecture, allowing the model to focus on different parts of the input data when generating output [11]. Tokenization, the process of breaking down input data into individual tokens, is also essential in Transformers, as it allows the model to process and understand the input data at a granular level. Different types of attention mechanisms, such as self-attention, multi-head attention, and hierarchical attention, can be used in Transformers to capture different types of relationships in the data [12]. Similarly, different tokenization techniques, such as word-level tokenization, subword-level tokenization, and character-level tokenization, can be used to capture different levels of granularity in the input data. The choice of attention mechanism and tokenization technique can significantly impact the performance of the Transformer model, and selecting the right combination is critical for achieving optimal results [13].

This study introduces a novel approach to bearing fault detection, leveraging a Transformer-based framework enhanced with a Temporal Decomposition Attention (TDA) mechanism and the innovative Hull Exponential Moving Average (HEMA). The core contributions of this method are summarized below:

- **Decomposition of Data**

Separating time series data into its seasonal and trend components enhances the understanding of underlying patterns and improves the fault detection process. By isolating the trend, the model captures long-term changes, such as gradual wear or drift in system performance. Simultaneously, the seasonal component highlights repetitive or cyclical behaviors, such as vibrations tied to operational cycles. This decomposition simplifies the data, removing systematic patterns and leaving residuals that are more indicative of potential anomalies or faults. Such an approach not only increases accuracy but also provides better interpretability, making it easier to pinpoint and address the root causes of detected faults.

- **Feature Extraction with HEMA**

The HEMA further strengthens the framework by effectively extracting features from the residual data after removing systematic patterns. HEMA reduces lag and noise by combining non-linear weighting and exponential smoothing, offering superior responsiveness compared to traditional moving averages. Its ability to adapt to changing patterns makes it an effective tool for identifying subtle deviations indicative of faults.

- **TDA Mechanism in Transformer**

The TDA mechanism enhances fault detection by enabling the model to flexibly adjust its focus across different time scales, effectively separating and attending to trend and seasonal components of the data. This approach captures both long-term dependencies and short-term periodic fluctuations with greater accuracy. Additionally, the explicit decomposition of data improves the interpretability of the model by clarifying which patterns influence predictions, making it especially valuable for applications requiring transparent and insightful decision-making in time series analysis. By integrating the TDA mechanism into the Transformer architecture, the model achieves a groundbreaking ability to capture complex temporal dynamics in comparison with traditional attention mechanisms.

2) **Bearing Fault Detection**

Fault detection is critical for maintaining the efficiency and safety of industrial operations, as it helps identify issues early and prevents costly failures. The CWRU dataset is a prominent benchmark for assessing fault detection techniques. This section provides an overview of current research on fault detection and presents the CWRU dataset as the foundation for the proposed approach.

**2.1) Literature Survey**

Deep learning methods have gained significant attention in the field of fault detection due to their ability to automatically extract complex features and patterns from large datasets, enhancing detection accuracy in industrial systems. Starnet is a machine learning-based framework that leverages probabilistic modeling and anomaly detection techniques to evaluate sensor trustworthiness and identify outliers in edge autonomy systems, enabling robust and reliable decision-making [14]. This research [15] utilizes deep learning techniques to predict financial market sequences, aiming to provide policymakers with accurate and timely insights to inform and enhance economic policies. The paper [16] demonstrates that gradient-based interpretation schemes of neural networks are vulnerable to universal adversarial perturbations (UPIs), which can significantly alter the interpretation of neural networks across various input samples without requiring knowledge of the specific input data.

Key approaches include CNNs, which are effective for spatial data and image-based fault detection; RNNs, particularly suited for time-series and sequential data; and autoencoders, which can learn compressed representations and detect anomalies by identifying deviations from learned patterns. Variants like Long Short-Term Memory (LSTM) networks are especially useful for capturing temporal dependencies in machine sensor data, while hybrid models combining CNNs with LSTM layers offer powerful solutions for complex fault detection tasks. Additionally, attention mechanisms have been integrated into deep learning models to improve their focus on relevant features, leading to improved fault diagnosis and localization accuracy. [17] provides a comprehensive review of the application of RNNs in mechanical fault diagnosis, summarizing the current state of research, highlighting key challenges and opportunities, and discussing the potential of RNNs in improving fault detection and diagnosis accuracy in mechanical systems. [18] presents a fault diagnosis approach for rotating machinery using RNNs, demonstrating the effectiveness of RNNs in learning temporal patterns and relationships in vibration signals to accurately identify and classify faults in rotating machinery. [19] proposes a novel fault diagnosis method that combines Convolutional Neural Networks (CNNs) and Long Short-Term Memory (LSTM) networks to effectively extract spatial and temporal features from complex system data, achieving improved fault diagnosis accuracy and robustness in identifying anomalies in complex systems.

Transformers have recently emerged as a promising approach for fault detection, particularly in scenarios involving sequential or time-series data, where their ability to capture long-range dependencies and model complex relationships is highly beneficial. In fault detection, Transformers can efficiently handle large-scale industrial sensor data, identifying patterns or anomalies that might indicate equipment failure. Variations of the standard Transformer architecture, such as the Vision Transformer (ViT) for image-based fault detection or Temporal Convolutional Networks (TCNs) combined with Transformers [20], have also been proposed to improve performance on fault classification and detection tasks. [21] proposes a novel Transformer-based neural network architecture for detecting and classifying faults in photovoltaic modules, leveraging its self-attention mechanism to effectively identify anomalies and faults in PV systems. [22] presents a novel variational attention-based Transformer network that provides interpretable results for rotary machine fault diagnosis, enabling the identification of key features and fault patterns while improving diagnostic accuracy and reliability. [23] presents a novel deep learning approach that combines CNNs with Transformer models to effectively detect faults in power system networks, leveraging the strengths of both architectures to improve accuracy and robustness in fault identification. [24] proposes a novel fault diagnosis method for planetary gearboxes that integrates time-series imaging feature fusion with a Transformer model, enabling the effective extraction and analysis of complex fault patterns and achieving improved diagnosis accuracy and reliability.

Based on our literature survey, it can be concluded that Transformers have surpassed traditional RNNs like LSTMs in sequence modeling tasks. They use self-attention mechanisms to capture complex patterns and relationships in parallel. This allows them to scale more easily to longer sequences and larger models. Transformers are more interpretable and require less training data than LSTMs. They have become the de facto standard for many NLP tasks

**2.2) Case Western Reserve University (CWRU) Dataset**

The CWRU dataset is one of the most widely used and benchmarked datasets in the field of bearing fault detection, providing a valuable resource for developing and testing fault detection algorithms, especially in the context of rotating machinery. This dataset was created by the center for Intelligent Maintenance Systems (IMS) at CWRU and is specifically designed to assist in the development of diagnostic models for detecting faults in bearings, a crucial component in many industrial systems [25]. The CWRU dataset has been widely cited in research papers and is considered a standard benchmark in the field. It has enabled the development of robust models for early fault detection, fault classification, and condition monitoring, helping industries reduce downtime and prevent catastrophic equipment failures. Its availability has also fostered collaboration across research institutions and companies to develop more accurate and efficient fault detection techniques [26].

Defects were introduced into the test bearings using Electrical Discharge Machining (EDM), ensuring precise and reproducible fault conditions. Defects vary by diameter and location within the bearing, simulating different types of common failures. The specific defect diameters include:

- Small: 0.007 inches (0.178 mm)
- Medium: 0.014 inches (0.356 mm)
- Large: 0.021 inches (0.533 mm)

Faults were introduced in three specific locations to cover various common bearing failure modes:

- Ball Fault: Located on the rolling element.
- Inner Race Fault: Positioned on the inner ring of the bearing.
- Outer Race Fault: Positioned on the outer ring of the bearing.

These controlled defects allow for reproducible, consistent fault conditions across the test bench, providing a robust dataset for fault detection and diagnostic model training. The motor operated under a controlled load condition with a 1 HP load applied and a fixed shaft speed of 1772 rpm, ensuring a consistent operational environment. Each accelerometer recorded vibration data at a high sampling frequency of 48 kHz, allowing for the capture of detailed time-series data that includes all relevant frequency components associated with bearing faults [27].

3) **Methodology**

This section outlines the key components and the proposed framework for bearing fault detection. It begins with an explanation of the HEMA for feature extraction, followed by a detailed description of the TDA mechanism. Finally, the integration of these components into the proposed HEMA-Transformer-TDA model is presented, highlighting the innovative approach and its design to address complex temporal patterns in fault detection tasks.

### 3.1) HEMA

The Hull Moving Average (HMA) is a technical indicator created to reduce the lag inherent in traditional moving averages and provide a more adaptive measure of price trends. Developed by Alan Hull in 2005, it uses a two-step weighted moving average (WMA) process [28]. First, a WMA is calculated on the price series, and then it is smoothed further using another WMA. The resulting value is determined by calculating the square root of the sum of the squares of the differences between the two WMAs. This non-linear weighting scheme adapts to market changes, allowing it to capture the underlying trend more accurately. The HMA is intended to outperform traditional moving averages by mitigating lag and noise.

$$WMA(n) = \frac{(n \times x_n + (n-1) \times x_{n-1} + \ldots + 1 \times x_1)}{(n + (n-1) + \ldots + 1)} \quad (1)$$

where $WMA$ is the weighted moving average, $n$ is the number of periods, $x_n, x_{n-1}, \ldots, x_1$ are the price data points. The first WMA, $WMA_1(n)$, is calculated using a window size of $\frac{n}{2}$:

$$WMA_1(n) = WMA\left(\frac{n}{2}\right), \quad (2)$$

Similarly, the second WMA, $WMA_2$, is calculated using the same window size:

$$WMA_2(n) = WMA\left(\frac{n}{2}\right), \quad (3)$$

The difference between the two WMAs, $Diff$, is computed as:

$$Diff = 2 \times WMA_1 - WMA_2, \quad (4)$$

Finally, the HMA is calculated using the $Diff$ value and a window size of $n$:

$$HMA = WMA(n, Diff) \quad (5)$$

Traditional moving averages, such as the Simple Moving Average (SMA), suffer from two major limitations: lag, which causes them to trail the data, and over-smoothing, which may obscure important trends. The HMA addresses these issues by reducing lag and enhancing responsiveness to market changes.

In comparison, the Exponential Moving Average (EMA) gives more weight to recent data, making it more responsive to changes in the market than the WMA. Unlike the WMA, which assigns a fixed weight to each data point, the EMA uses an exponential decay factor to place more emphasis on recent data points. This characteristic allows the EMA to adapt quickly to market shifts, making it more sensitive to recent price movements and less prone to lag. Furthermore, the EMA is less affected by outliers, making it a more reliable choice for traders.

$$EMA = (\alpha \times x_n) + \left((1-\alpha) \times EMA_{(n-1)}\right) \tag{6}$$

where $\alpha$ is the smoothing factor, and typically it is 0.2. The first and second EMAs, $EMA_1(n)$ and $EMA_2(n)$, are calculated using a window size of $\frac{n}{2}$:

$$EMA_1(n) = EMA\left(\frac{n}{2}\right), \tag{7}$$

$$EMA_2(n) = EMA\left(\frac{n}{2}\right), \tag{8}$$

The difference $Diff$ is then calculated as:

$$Diff = 2 \times EMA_1 - EMA_2, \tag{9}$$

Finally, the HEMA is calculated using $Diff$:

$$HEMA = EMA(n, Diff) \tag{10}$$

The novel HEMA addresses the key issues of the traditional WMA by combining the responsiveness of the EMA and the reduced lag of the HMA. Unlike the WMA, which suffers from lag and over-smoothing, the HEMA is more adaptive, giving greater weight to recent data and capturing trends more accurately without losing important information. This makes it a superior choice for time series prediction, fault detection, and financial analysis.

**3.2) TDA mechanism**

The attention mechanism is a fundamental component of modern deep learning models, especially in sequence-based tasks like natural language processing and time series analysis. It allows the model to dynamically weigh and focus on different parts of the input sequence when producing the output. This selective attention mechanism is particularly effective because it enables the model to prioritize important information while disregarding less relevant data, which improves performance, especially in tasks involving long-range dependencies or complex relationships between elements in the sequence.

$$Attention(Q, K, V) = Softmax\left(\frac{Q \cdot K^T}{\sqrt{d_k}}\right) \times V. \tag{11}$$

where Queries ($Q$) represent the information we seek from the input sequence, Keys ($K$) provide context about other elements in the sequence, and Values (V) contain the actual content associated with each element, $d_k$ is the dimensionality of the keys, and the scaling factor $\sqrt{d_k}$ prevents large dot-product values, stabilizing the training process.

A novel attention mechanism, the TDA, is introduced to enhance the Transformer model's ability to capture both trend-based and seasonally-based temporal dependencies in time series data. This mechanism incorporates Temporal Bias Encoding to separate and effectively model the trend and seasonal components of the time series. The attention mechanism for the trend and season component are formulated as follows:

$$Attention_{trend}(Q, K, V_{trend}, \alpha_t^{trend}) = Softmax\left(\frac{(Q \cdot K^T) \cdot \alpha_t^{trend}}{\sqrt{d_k}}\right) \times V_{trend} \quad (12)$$

$$Attention_{season}(Q, K, V_{season}, \alpha_t^{season}) = Softmax\left(\frac{(Q \cdot K^T) \cdot \alpha_t^{season}}{\sqrt{d_k}}\right) \times V_{season} \quad (13)$$

where $V_{trend}$ and $V_{season}$ represent the trend component of the value matrix, while $\alpha_t^{trend}$ and $\alpha_t^{season}$ are learnable temporal decay factor that adjusts the emphasis on the trend component over time. The term $d_k$ refers to the dimensionality of the key vector. The final output is obtained by combining the outputs of the trend and seasonal attentions:

$$Attention = Attention_{trend}(Q, K, V_{trend}) + Attention_{season}(Q, K, V_{season}) \quad (14)$$

This formulation allows the model to capture long-term dependencies in the trend component by modulating attention based on the temporal bias encoded in $\alpha_t^{trend}$, while also capturing short-term periodic dependencies in the seasonal component through $\alpha_t^{season}$. By separately modeling these two components, the approach enhances the model's ability to handle complex time series data with both trend and seasonal patterns.

**3.3) Proposed Methodology**

In the proposed method, the trend and seasonality components of the CWRU dataset are first identified and removed to isolate residual patterns. This step is crucial as it eliminates predictable, systematic variations that are inherent to the data but not indicative of faults, such as gradual drifts (trend) or repeating periodic patterns (seasonality). By doing so, the remaining residual patterns emphasize deviations or noise that are more likely linked to anomalies or faults, making them more suitable for diagnostic analysis.

Once the residual data is isolated, feature extraction methods are applied to capture its key characteristics. The HEMA is employed to smooth the data while retaining responsiveness to recent changes, effectively highlighting significant deviations that might indicate faults. Additionally, statistical measures like skewness and kurtosis are calculated. Skewness measures the asymmetry of the data distribution, which could reveal irregular shifts, while kurtosis assesses the sharpness of data peaks, which might indicate rare or extreme events. Together, these features provide a comprehensive understanding of the residual data's behavior.

To enhance the analysis further, the TDA mechanism, integrated within the Transformer architecture, is applied. This mechanism explicitly focuses on modeling trend-based and seasonally-based temporal dependencies in the dataset. By decomposing the Transformer's attention mechanism into distinct components for trends and seasonality, TDA effectively captures both long-term dependencies, such as persistent drifts in operational behavior, and short-term periodic fluctuations, like regular operational cycles. This dual-focus design enables the Transformer to adapt to the data's underlying temporal structure while improving its ability to identify anomalies with greater precision.

In combination, these processes provide a robust framework for fault detection, integrating statistical insights, adaptive smoothing, and temporal modeling to handle complex time series data effectively.

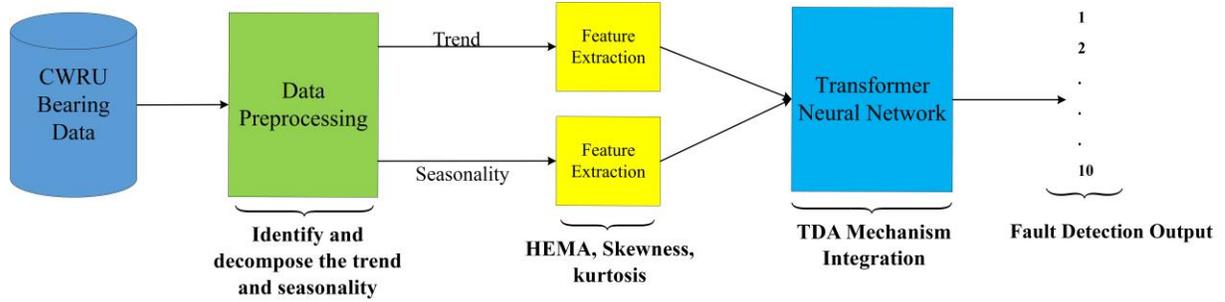

Fig. 1. Proposed Bearing Fault Detection Process.

**4) Performance Evaluation**

This section assesses the performance of the proposed models through accuracy analysis, confusion matrices, and essential fault detection metrics. Each subsection offers a detailed evaluation of the models' efficiency, dependability, and capability to handle the challenges of bearing fault detection. Table 1 outlines the specifics of the faults in the CWRU dataset.

Table 1. Fault Types and Severity Levels in the CWRU Bearing Dataset.

| Fault Type | Severity | Description | Abbreviation |
|---|---|---|---|
| Inner Race | Small | Fault in the inner race of the bearing | IR_007_1 |
| Inner Race | Medium | Fault in the inner race of the bearing | IR_014_1 |
| Inner Race | Large | Fault in the inner race of the bearing | IR_021_1 |
| Outer Race | Small | Fault in the outer race of the bearing | OR_007_6_1 |
| Outer Race | Medium | Fault in the outer race of the bearing | OR_014_6_1 |
| Outer Race | Large | Fault in the outer race of the bearing | OR_021_6_1 |
| Ball | Small | Fault in the balls of the bearing | Ball_007_1 |
| Ball | Medium | Fault in the balls of the bearing | Ball_014_1 |
| Ball | Large | Fault in the balls of the bearing | Ball_021_1 |

## 4.1) Accuracy Analysis for Bearing Fault Detection Models

Accuracy is a performance metric that measures the proportion of correct predictions made by the model compared to the total number of predictions, as follows:

$$Accuracy = \frac{TP + TN}{TP + FN + TN + FP} \tag{15}$$

Figure 4 illustrates the fault classification accuracy of various machine learning models applied to the CWRU dataset for bearing fault detection. The models tested include AlexNet, GoogleNet, ResNet, Wavelet-Attention, Transformer, HEMA-Transformer, and HEMA-Transformer-TDA.

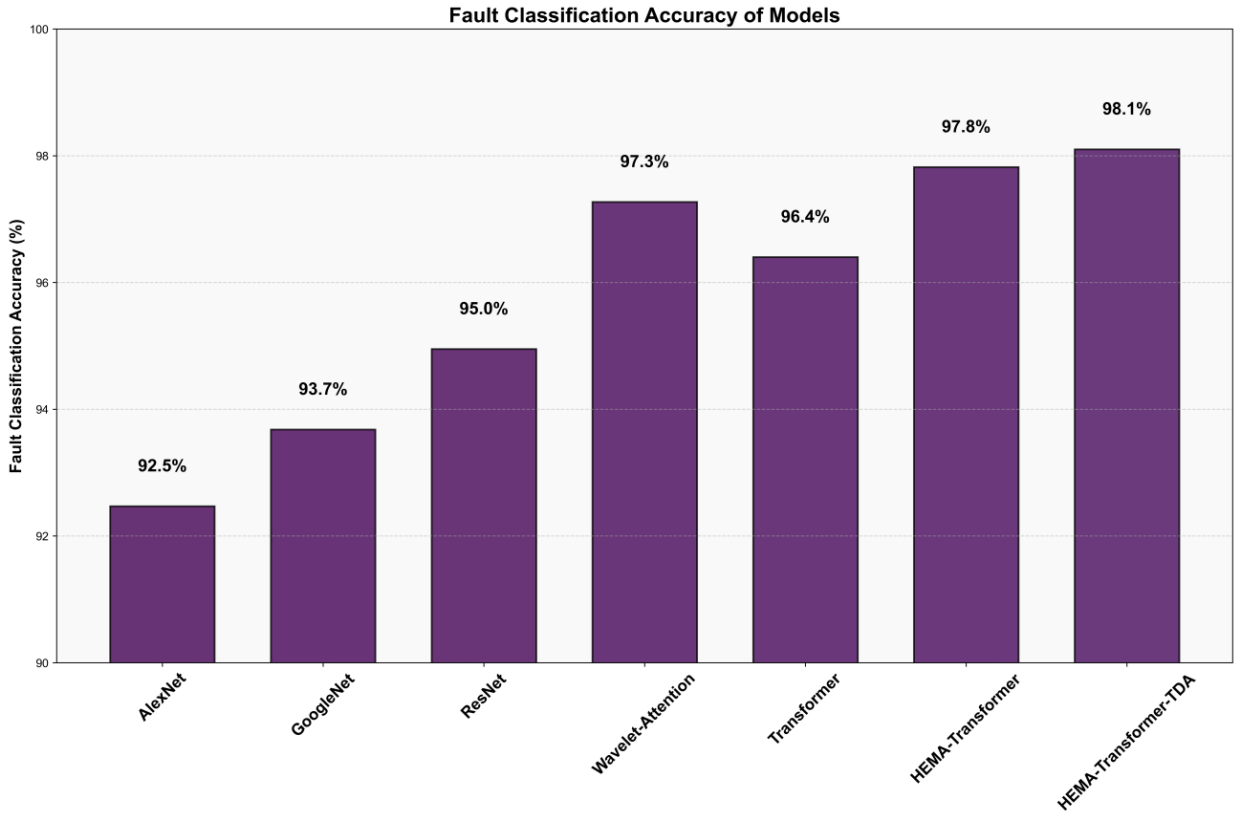

Fig. 2. Accuracy of the bearing fault detection methods using AlexNet [29], GoogleNet [30], ResNet [31], Wavelet-Attention[32], Transformer, HEMA-Transformer, HEMA-Transformer-TDA

According to Fig. 2, The results demonstrate that the HEMA-Transformer-TDA model achieves the highest accuracy of 98.1%, surpassing all other models tested. Fig. 4 provides compelling evidence that the HEMA-Transformer-TDA model offers a promising approach for accurate bearing fault detection. The combination of advanced signal processing techniques, deep learning architectures, and attention mechanisms tailored for temporal dependencies appears to be highly effective in extracting relevant features and classifying faults with high precision. Here are the key observations:

- The Transformer model shows a notable improvement over traditional CNNs like AlexNet [29], GoogleNet [30], and ResNet [31], highlighting the effectiveness of attention mechanisms for capturing temporal dependencies in time series data.
- The incorporation of the HEMA filter into the Transformer model (HEMA-Transformer) further boosts accuracy, indicating the beneficial role of pre-processing techniques in isolating fault-related features.
- The addition of the TDA mechanism to the HEMA-Transformer model yields the highest accuracy, suggesting that explicitly modeling trend-based and seasonally-based dependencies within the attention mechanism is crucial for robust fault detection.

**4.2) Confusion Matrix**

The confusion matrix is used to evaluate the performance of the fault detection method by showing the distribution of predicted versus actual class labels, highlighting the accuracy and misclassification rates across different fault categories. Figures 3, 4, and 5 present the confusion matrices for the Transformer, HEMA-Transformer, and HEMA-Transformer-TDA models, respectively, illustrating their classification performance across various fault classes.

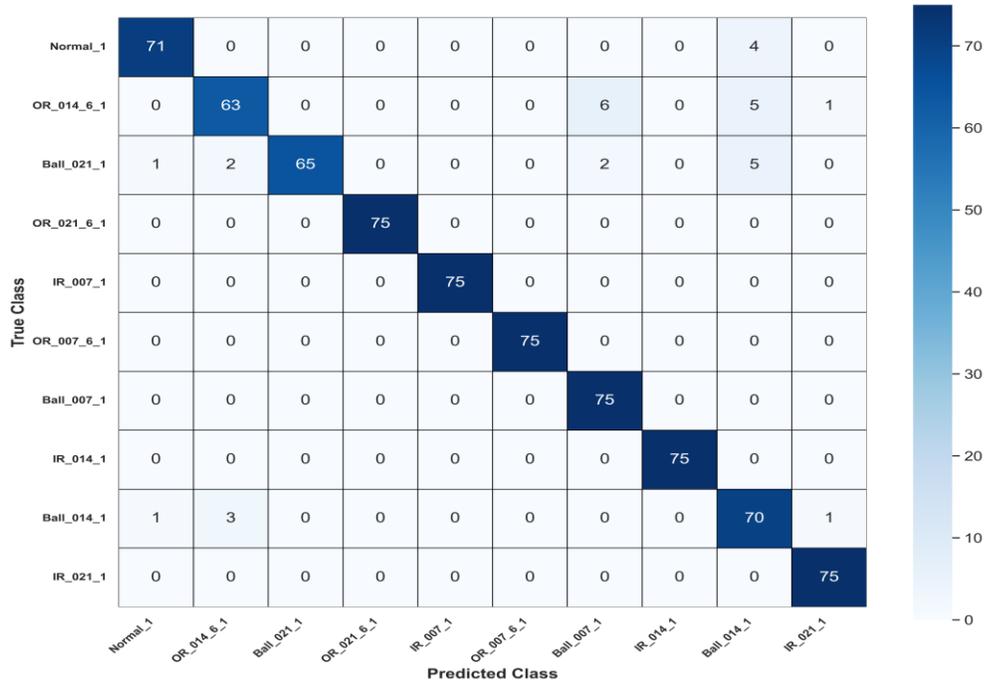

Fig. 3. Confusion Matrix of the bearing fault detection using the Transformer

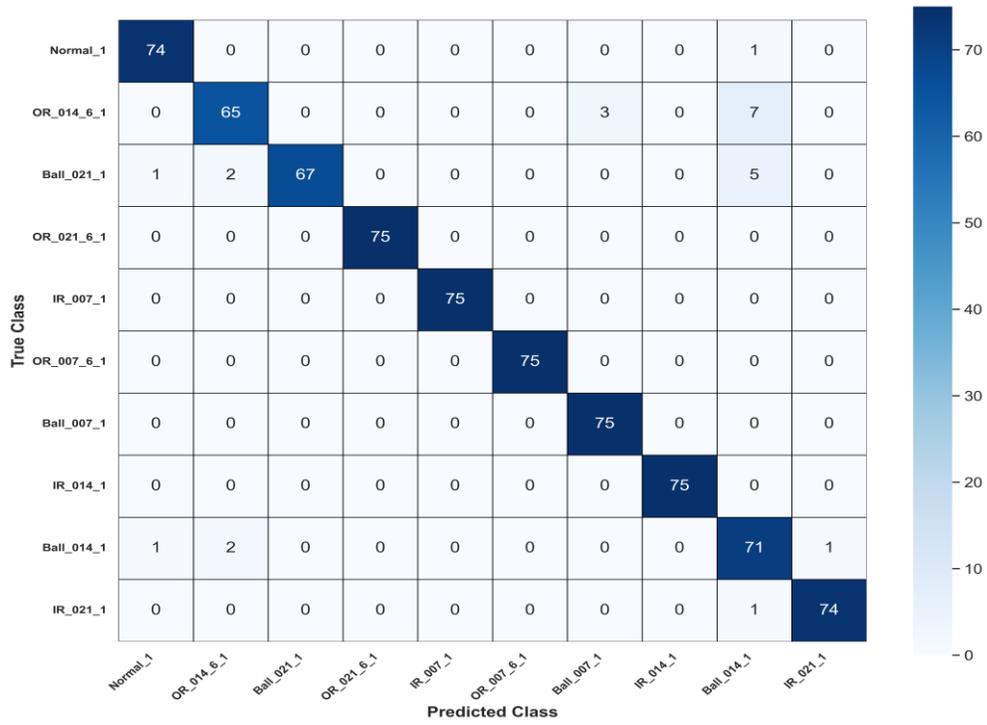

Fig. 4. Confusion Matrix of the bearing fault detection using the HEMA-Transformer

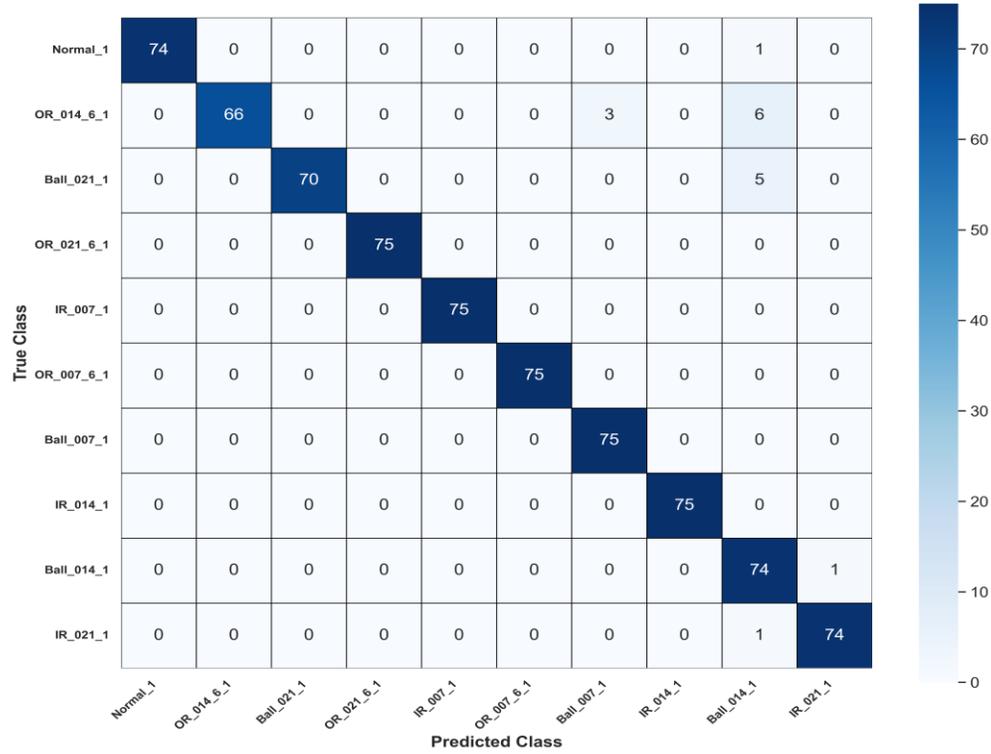

Fig. 5. Confusion Matrix of the bearing fault detection using the HEMA-Transformer-TDA

The HEMA-Transformer-TDA model exhibits superior performance compared to the base Transformer model across multiple fault classes. Notably, the HEMA-Transformer achieves the highest performance for the **Normal_1** class, with 74 correctly classified instances. Moreover, misclassification rates are significantly reduced for challenging fault classes such as **Ball_014_1** and **IR_021_1** in both the HEMA-Transformer and HEMA-Transformer-TDA models, reflecting their enhanced accuracy in fault detection. These improvements highlight the effectiveness of HEMA-based models in addressing the complexities of time-series fault classification.

The incorporation of temporal dependency modeling through the TDA module further enhances the HEMA-Transformer-TDA's performance by producing balanced classification results across all classes. This mechanism captures intricate time-series patterns, effectively isolating long-term dependencies and periodic fluctuations. Additionally, the model's integration of trend and seasonal decomposition improves interpretability by facilitating the detection of distinct temporal patterns, particularly for faults with fluctuating behaviors like **IR_021_1**. This enhanced interpretability is instrumental in understanding fault progression and the underlying causes of failures.

**4.3) Bearing Fault Detection Metrics**

In this subsection, an in-depth analysis is conducted on the performance metrics used to evaluate the effectiveness of the proposed fault detection approach. A comprehensive evaluation of the method's performance is conducted using a range of key metrics, including Precision, Recall, F1-score, False Alarm Rate (FAR), and Missed Alarm Rate (MAR). These metrics are employed to provide detailed insights into the method's ability to accurately identify and classify faults.

- Precision:

The proportion of true positive predictions out of all positive predictions made by the model. It measures the accuracy of positive predictions.

$$Precision = \frac{TP}{TP + FP}. \tag{16}$$

- Recall

The proportion of true positive predictions out of all actual positive instances in the dataset. It indicates the model's ability to detect all positive instances.

$$Recall = \frac{TP}{TP + FN}. \tag{17}$$

- F1-score

The harmonic mean of Precision and Recall. It provides a single metric that balances both false positives and false negatives.

$$F1-score = 2\frac{Precision \times Recall}{Precision + Recall}. \tag{18}$$

- FAR

The proportion of false positives (incorrectly predicted positives) out of all actual negatives. It measures how often the model falsely predicts a positive result.

$$FAR = \frac{FP}{TN + FP}. \tag{19}$$

- MAR

The proportion of false negatives (incorrectly predicted negatives) out of all actual positives. It measures how often the model fails to detect positive instances.

$$MAR = \frac{FN}{FN + TP}. \tag{20}$$

Figs. (6-10) illustrate the fault detection metrics.

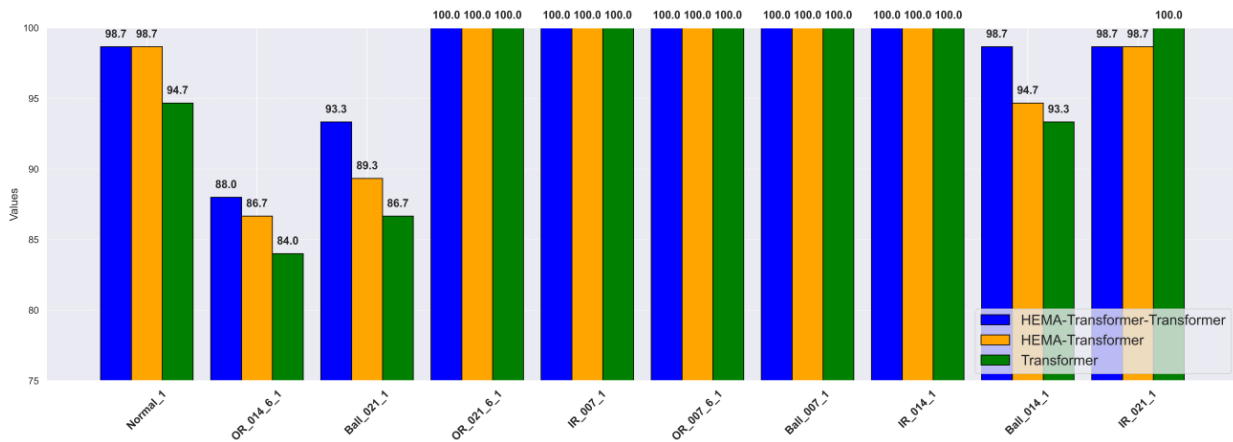

Fig. 6. Recall of using Transformer, HEMA-Transformer, HEMA-Transformer-TDA methods.

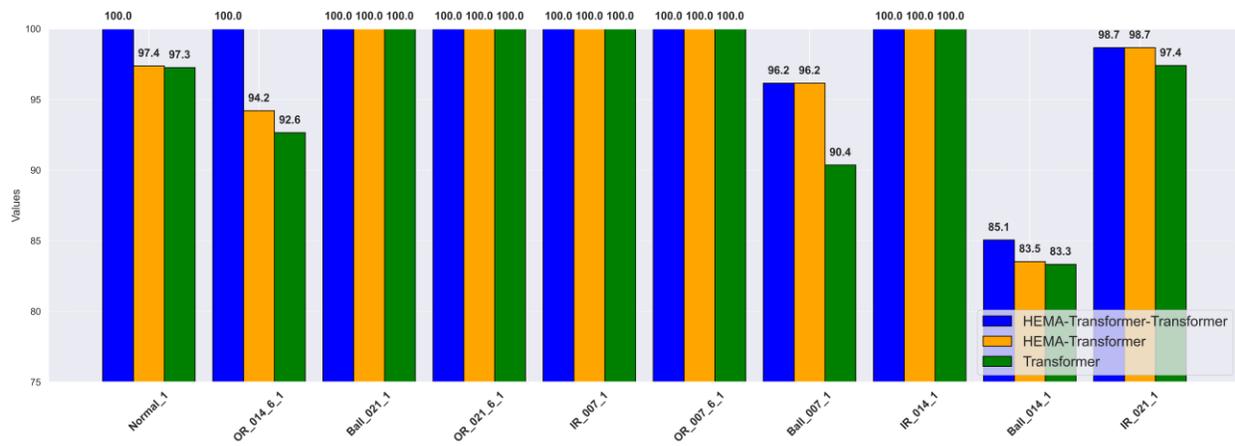

Fig. 7. Precision of using Transformer, HEMA-Transformer, HEMA-Transformer-TDA methods.

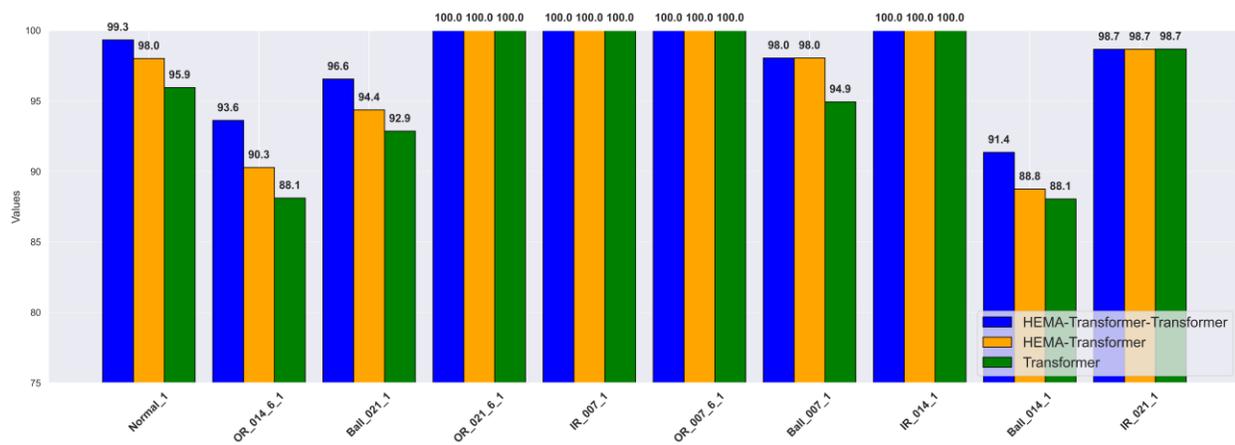

Fig. 8. F1-score of using Transformer, HEMA-Transformer, HEMA-Transformer-TDA methods.

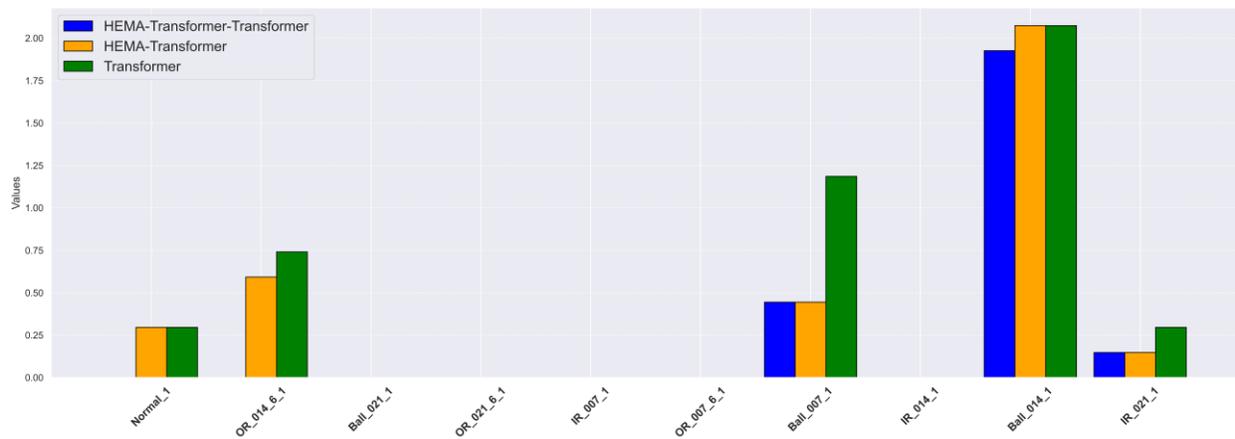

Fig. 9. FAR of using Transformer, HEMA-Transformer, HEMA-Transformer-TDA methods.

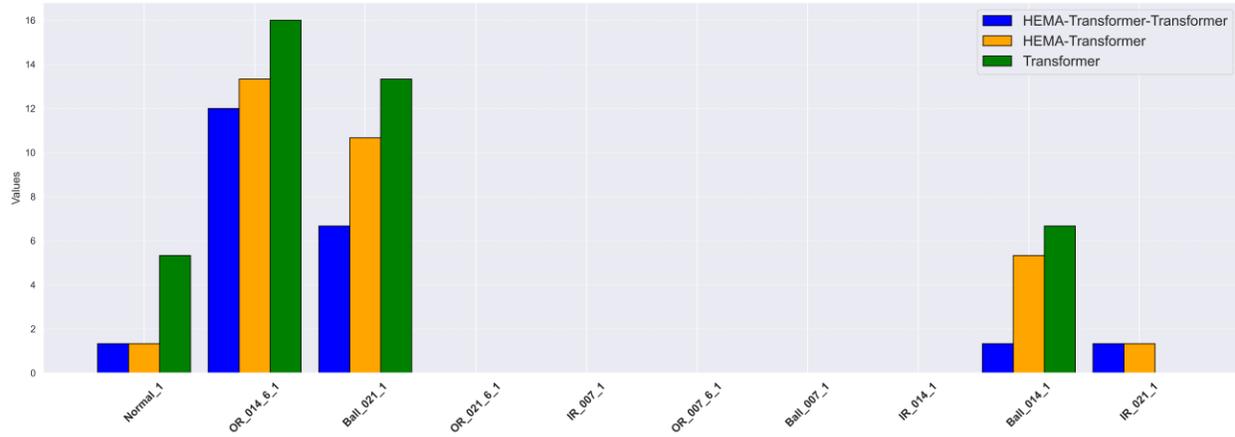

Fig. 10. MAR of using Transformer, HEMA-Transformer, HEMA-Transformer-TDA methods.

Based on the (6-10), it can be concluded that:

- **Precision**: Across all models, precision values are exceptionally high (close to 1) for most classes, indicating that the models perform well in identifying true positive instances and avoiding false positives. However, precision slightly drops for classes like **Ball_014_1** (0.85), which might be due to more challenging characteristics of this fault type that make it harder to differentiate from others.

- **Recall**: Recall values are consistently high for all classes in all three models. The **IR_007_1** and **IR_021_1** classes achieve perfect recall (1.0), meaning that these fault types are detected without missing any true instances. However, some classes like **Ball_014_1** and **Ball_007_1** have slightly lower recall (around 0.88–0.93), suggesting that the model might occasionally miss detecting some faults of these types.

- **F1-score**: The F1-scores are high across all models, indicating strong performance in terms of both precision and recall. The HEMA-Transformer-TDA model consistently achieves an F1-score of 1 for several classes, indicating excellent balanced performance. For other classes like **Ball_014_1**, the F1-scores are lower, but still strong, especially for challenging classes where precision and recall are more balanced.

- **FAR**: All models show very low FAR values (mostly 0), which indicates that the models are extremely effective in avoiding false alarms. The **Ball_007_1** and **IR_014_1** classes show a slightly higher FAR, but still, the values remain small. This suggests that the models are good at distinguishing between faults and non-fault conditions.

- **MAR**: The MAR is also low across all models, with a few exceptions. The classes **Ball_014_1** and **IR_007_1** show slightly higher MAR values (around 0.12–0.13), indicating that a small portion of the faults in these classes are missed. However, the MAR values remain relatively small, suggesting that missed detections are not a major issue.

**5) Conclusion**

This study presents a novel approach for bearing fault detection in rotating machinery, combining a Transformer neural network with TDA and HEMA feature extraction. The TDA mechanism effectively captures both long-term trends and seasonal patterns within the time series data, enabling the model to learn intricate temporal dependencies and improve fault classification accuracy. Our experimental results on the CWRU dataset demonstrate that the HEMA-Transformer-TDA model significantly outperforms traditional Transformer models and other state-of-the-art methods. The model achieves a high overall accuracy of 98.1%, with excellent precision, recall, and F1-scores for most fault classes. Furthermore, the model exhibits low FAR and MAR, indicating a high degree of reliability and robustness in fault detection. The incorporation of TDA enhances the model's interpretability by providing insights into the temporal dynamics of the fault signals. This enables a deeper understanding of fault progression and facilitates more informed maintenance decisions.

These findings emphasize the significance of combining advanced attention mechanisms with effective feature extraction techniques for fault diagnosis. The HEMA-Transformer-TDA model not only enhances accuracy but also provides improved interpretability, making it a promising solution for predictive maintenance in industrial applications. Future research could explore the applicability of this approach to other time-series datasets with complex temporal patterns, further expanding its utility in fault detection and time-series analysis tasks.

**Funding**

This research received no external funding.

**Conflicts of Interest**

The authors declare no conflict of interest.

## 6) Reference

[1]   J. Hendriks, P. Dumond, and D. A. Knox, "Towards better benchmarking using the CWRU bearing fault dataset," *Mech. Syst. Signal Process.*, vol. 169, p. 108732, 2022.

[2]   D. Neupane and J. Seok, "Bearing fault detection and diagnosis using case western reserve university dataset with deep learning approaches: A review," *Ieee Access*, vol. 8, pp. 93155–93178, 2020.

[3]   C. Herff and D. J. Krusienski, "Extracting features from time series," *Fundam. Clin. data Sci.*, pp. 85–100, 2019.

[4]   L. Zhang and N. Hu, "Time domain synchronous moving average and its application to gear fault detection," *IEEE Access*, vol. 7, pp. 93035–93048, 2019.

[5]     A. Khorram, M. Khalooei, and M. Rezghi, "End-to-end CNN+ LSTM deep learning approach for bearing fault diagnosis," *Appl. Intell.*, vol. 51, no. 2, pp. 736–751, 2021.

[6]     F. Dao, Y. Zeng, and J. Qian, "Fault diagnosis of hydro-turbine via the incorporation of bayesian algorithm optimized CNN-LSTM neural network," *Energy*, vol. 290, p. 130326, 2024.

[7]     W. Li *et al.*, "Fault diagnosis using variational autoencoder GAN and focal loss CNN under unbalanced data," *Struct. Heal. Monit.*, p. 14759217241254120, 2024.

[8]     J. Zhao, X. Feng, J. Wang, Y. Lian, M. Ouyang, and A. F. Burke, "Battery fault diagnosis and failure prognosis for electric vehicles using spatio-temporal transformer networks," *Appl. Energy*, vol. 352, p. 121949, 2023.

[9]     J. Li, Y. Bao, W. Liu, P. Ji, L. Wang, and Z. Wang, "Twins transformer: Cross-attention based two-branch transformer network for rotating bearing fault diagnosis," *Measurement*, vol. 223, p. 113687, 2023.

[10]    C. Chen, C. Liu, T. Wang, A. Zhang, W. Wu, and L. Cheng, "Compound fault diagnosis for industrial robots based on dual-transformer networks," *J. Manuf. Syst.*, vol. 66, pp. 163–178, 2023.

[11]    X. Li *et al.*, "Deep learning attention mechanism in medical image analysis: Basics and beyonds," *Int. J. Netw. Dyn. Intell.*, pp. 93–116, 2023.

[12]    S. Choo and W. Kim, "A study on the evaluation of tokenizer performance in natural language processing," *Appl. Artif. Intell.*, vol. 37, no. 1, p. 2175112, 2023.

[13]    M. L. Abimouloud, K. Bensid, M. Elleuch, O. Aiadi, and M. Kherallah, "Vision Transformer Based Tokenization for Enhanced Breast Cancer Histopathological Images Classification," in *IFIP International Conference on Artificial Intelligence Applications and Innovations*, Springer, 2024, pp. 255–267.

[14]    N. Darabi, S. Tayebati, S. Ravi, T. Tulabandhula, and A. R. Trivedi, "Starnet: Sensor trustworthiness

and anomaly recognition via approximated likelihood regret for robust edge autonomy," *arXiv Prepr. arXiv2309.11006*, 2023, doi: 10.48550/arXiv.2309.11006.

[15] S. Salahshour, M. Salimi, K. Tehranian, N. Erfanibehrouz, M. Ferrara, and A. Ahmadian, "Deep prediction on financial market sequence for enhancing economic policies," *Decis. Econ. Financ.*, pp. 1–20, 2024, doi: 10.1007/s10203-024-00488-4.

[16] H. E. Oskouie, L. Levine, and M. Sarrafzadeh, "Exploring Cross-model Neuronal Correlations in the Context of Predicting Model Performance and Generalizability," *arXiv Prepr. arXiv2408.08448*, 2024.

[17] J. Zhu, Q. Jiang, Y. Shen, C. Qian, F. Xu, and Q. Zhu, "Application of recurrent neural network to mechanical fault diagnosis: a review," *J. Mech. Sci. Technol.*, vol. 36, no. 2, pp. 527–542, 2022, doi: 10.1007/s12206-022-0102-1.

[18] Y. Zhang, T. Zhou, X. Huang, L. Cao, and Q. Zhou, "Fault diagnosis of rotating machinery based on recurrent neural networks," *Measurement*, vol. 171, p. 108774, 2021.

[19] T. Huang, Q. Zhang, X. Tang, S. Zhao, and X. Lu, "A novel fault diagnosis method based on CNN and LSTM and its application in fault diagnosis for complex systems," *Artif. Intell. Rev.*, vol. 55, no. 2, pp. 1289–1315, 2022.

[20] M. Xing, W. Ding, H. Li, and T. Zhang, "A power transformer fault prediction method through temporal convolutional network on dissolved gas chromatography data," *Secur. Commun. Networks*, vol. 2022, no. 1, p. 5357412, 2022.

[21] E. A. Ramadan, N. M. Moawad, B. A. Abouzalm, A. A. Sakr, W. F. Abouzaid, and G. M. El-Banby, "An innovative transformer neural network for fault detection and classification for photovoltaic modules," *Energy Convers. Manag.*, vol. 314, p. 118718, 2024.

[22] Y. Li, Z. Zhou, C. Sun, X. Chen, and R. Yan, "Variational attention-based interpretable transformer

network for rotary machine fault diagnosis," *IEEE Trans. neural networks Learn. Syst.*, 2022.

[23] J. B. Thomas, S. G. Chaudhari, K. V Shihabudheen, and N. K. Verma, "CNN-based transformer model for fault detection in power system networks," *IEEE Trans. Instrum. Meas.*, vol. 72, pp. 1–10, 2023.

[24] R. Wu, C. Liu, T. Han, J. Yao, and D. Jiang, "A planetary gearbox fault diagnosis method based on time-series imaging feature fusion and a transformer model," *Meas. Sci. Technol.*, vol. 34, no. 2, p. 24006, 2022.

[25] D. Neupane and J. Seok, "Bearing fault detection and diagnosis using case western reserve university dataset with deep learning approaches: A review," *IEEE Access*, vol. 8, pp. 93155–93178, 2020, doi: 10.1109/ACCESS.2020.2990528.

[26] X. Zhang, B. Zhao, and Y. Lin, "Machine learning based bearing fault diagnosis using the case western reserve university data: A review," *Ieee Access*, vol. 9, pp. 155598–155608, 2021.

[27] K. K. Raj, S. Kumar, R. R. Kumar, and M. Andriollo, "Enhanced Fault Detection in Bearings Using Machine Learning and Raw Accelerometer Data: A Case Study Using the Case Western Reserve University Dataset," *Information*, vol. 15, no. 5, p. 259, 2024.

[28] A. Hull, *Active Investing*. Australia, Wrightbooks, 2001.

[29] C. Alippi, S. Disabato, and M. Roveri, "Moving convolutional neural networks to embedded systems: the alexnet and VGG-16 case," in *2018 17th ACM/IEEE International Conference on Information Processing in Sensor Networks (IPSN)*, IEEE, 2018, pp. 212–223.

[30] G. Cao, K. Zhang, K. Zhou, H. Pan, Y. Xu, and J. Liu, "A feature transferring fault diagnosis based on WPDR, FSWT and GoogLeNet," in *2020 IEEE International Instrumentation and Measurement Technology Conference (I2MTC)*, IEEE, 2020, pp. 1–6.

[31] L. Wen, X. Li, and L. Gao, "A transfer convolutional neural network for fault diagnosis based on

ResNet-50," *Neural Comput. Appl.*, vol. 32, no. 10, pp. 6111–6124, 2020.

[32] Y. Yan, Q. Liu, and X. qin Gao, "Motor fault diagnosis algorithm based on wavelet and attention mechanism," *J. Sensors*, vol. 2021, no. 1, p. 3782446, 2021.